\documentclass[10pt,twocolumn,letterpaper]{article}

\usepackage{cvpr}
\usepackage{times}
\usepackage{epsfig}
\usepackage{graphicx}
\usepackage{amsmath}
\usepackage{amssymb}
 \usepackage{epsfig}
 \usepackage{latexsym}
 \usepackage{amsmath}
\usepackage{tabularx}

 \usepackage{multirow}
 \usepackage{stfloats}
 \usepackage{amssymb}
 \usepackage{color}
 \usepackage{subfig}
\usepackage{relsize}

\newcommand{\comment}[1]{}

    {\begin{Sbox}\begin{minipage}}%
    {\end{minipage}\end{Sbox}\fbox{\TheSbox}}

\title{Learning Gaze Transitions from Depth to Improve Video Saliency Estimation}

\author{G. Leifman$^{1}$, D. Rudoy, T. Swedish$^{1}$, E. Bayro-Corrochano and R. Raskar$^{1}$
\thanks{$^{1}$ MIT Media Lab, Massachusetts Institute of Technology}
}

\cvprfinalcopy 

\ifcvprfinal\fi

\begin{document}

\maketitle

\begin{abstract}
In this paper we introduce a novel Depth-Aware Video Saliency approach to predict human focus of attention when viewing RGBD videos on regular 2D screens.
We train a generative convolutional neural network which predicts a saliency map for a frame, given the fixation map of the previous frame.
Saliency estimation in this scenario is highly important since in the near future 3D video content will be easily acquired and yet hard to display.
This can be explained, on the one hand, by the dramatic improvement of 3D-capable acquisition equipment.
On the other hand, despite the considerable progress in 3D display technologies, most of the 3D displays are still expensive and require wearing special glasses.
To evaluate the performance of our approach, we present a new comprehensive database of eye-fixation ground-truth for RGBD videos.
Our experiments indicate that integrating depth into video saliency calculation is beneficial.
We demonstrate that our approach outperforms state-of-the-art methods for video saliency, achieving 15\% relative improvement.
\end{abstract}

\section{Introduction}
\label{sec:introduction}

In recent years we have witnessed a dramatic improvement of 3D-capable acquisition equipment; 3D cameras, \eg Kinect and RealSense, have become highly popular and affordable.
Moreover, in the near future many laptops and tablets are expected to be shipped with integrated 3D cameras.
We also see a considerable progress in 3D display technologies, \eg~\cite{hirsch2014compressive}.
However, high-quality 3D displays are still expensive and not easily accessible to the average consumer.
Combining the above two factors leads to a world where the 3D content is easy to acquire but hard to display.
Thus we predict human foci of attention when viewing 3D content on regular 2D screens.

Saliency detection in video sequences has attracted a lot of attention in recent years due to its contribution for various computer vision applications, which include segmentation, classification, key-frame selection, retargeting and compression.
3D visual information supplies a powerful cue for saliency analysis.
This has been shown by numerous studies that investigate the effect of depth information for image and video saliency~\cite{ciptadi2013depth,judepth2014ICIP,lang2012depth,niu2012leveraging,ouerhani2000computing}.
The eye movement patterns in 3D stereoscopic moving sequences have been recently investigated as well~\cite{huynh2011importance,huynh2011examination} and were proven to differ from the eye movement when viewing the same content on a 2D screen.
This difference is beyond the scope of this paper since we focus on the scenarios where depth information exists but is not displayed to the viewer.


\begin{figure}[t]
\centering
   \begin{tabular}{ccc}
   \multicolumn{3}{c}{\includegraphics[width=0.9\linewidth]{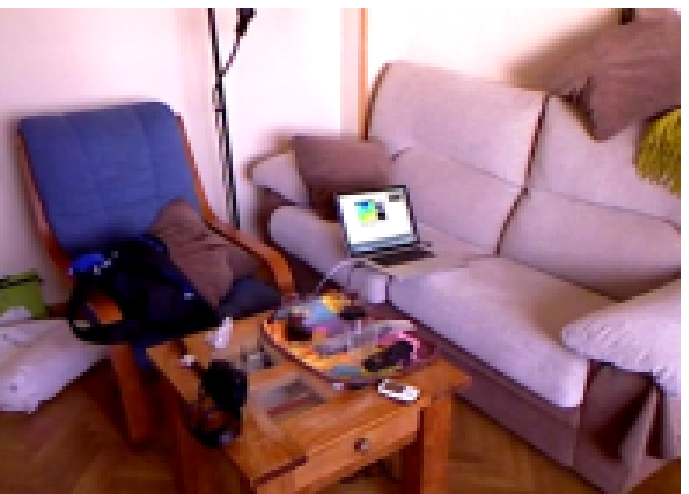} } \\
    {\small Ground-truth} & \hspace*{-0.1in} {\small Our approach} & \hspace*{-0.1in}{\small Rudoy et al.~\cite{rudoy2013Learning}} \\
       \includegraphics[width=0.28\linewidth]{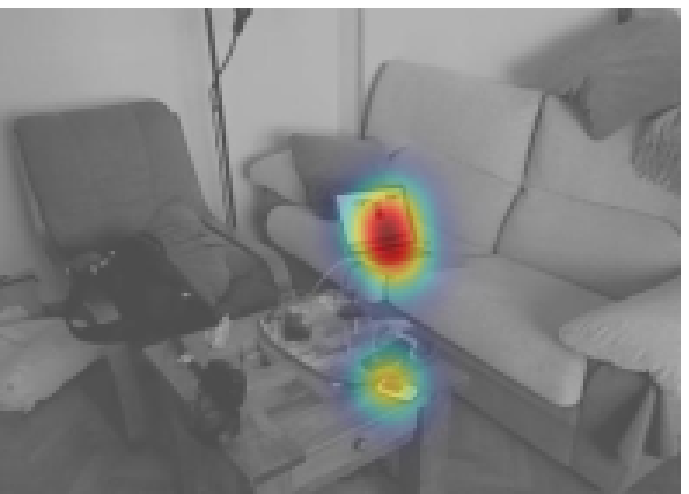} &
       \hspace*{-0.1in} \includegraphics[width=0.28\linewidth]{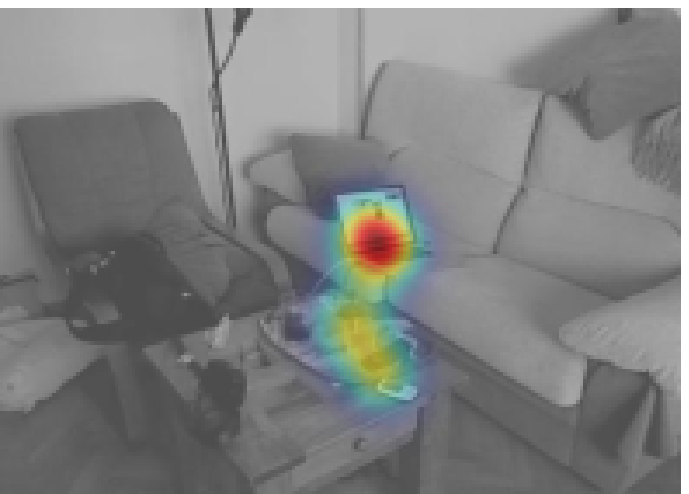} &
       \hspace*{-0.1in} \includegraphics[width=0.28\linewidth]{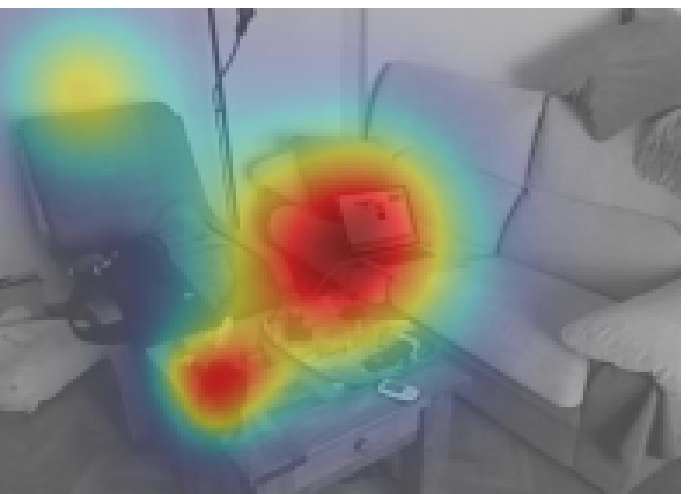}
   \end{tabular}
\caption{Our depth-aware video saliency is more similar to the ground-truth than the state-of-the-art method~\cite{rudoy2013Learning}.}
\label{Fig:teaser}
\end{figure}

\begin{figure}[t]
\centering
   \begin{tabular}{cc}
   \includegraphics[width=0.4\linewidth]{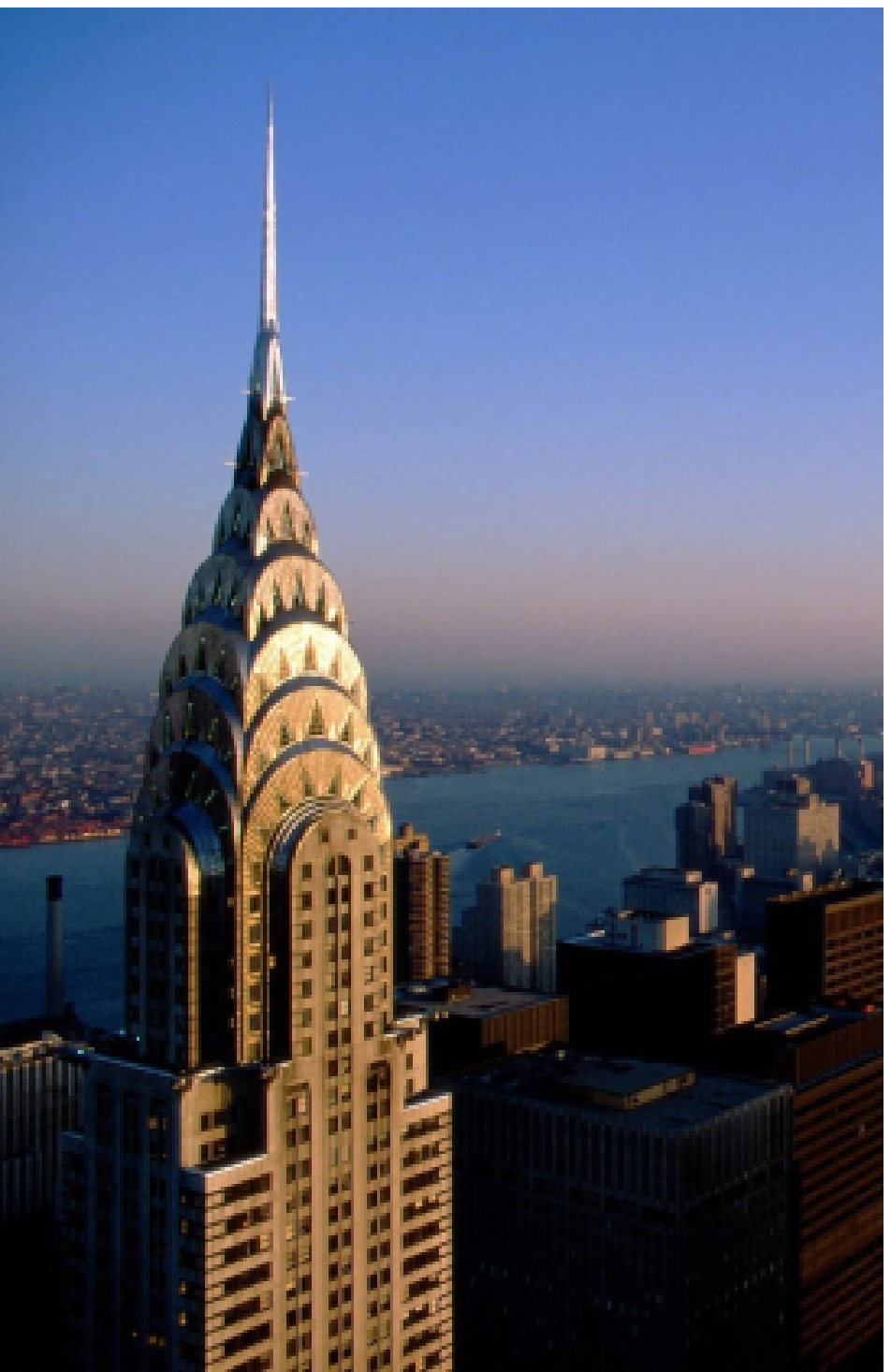} &
   \includegraphics[width=0.4\linewidth]{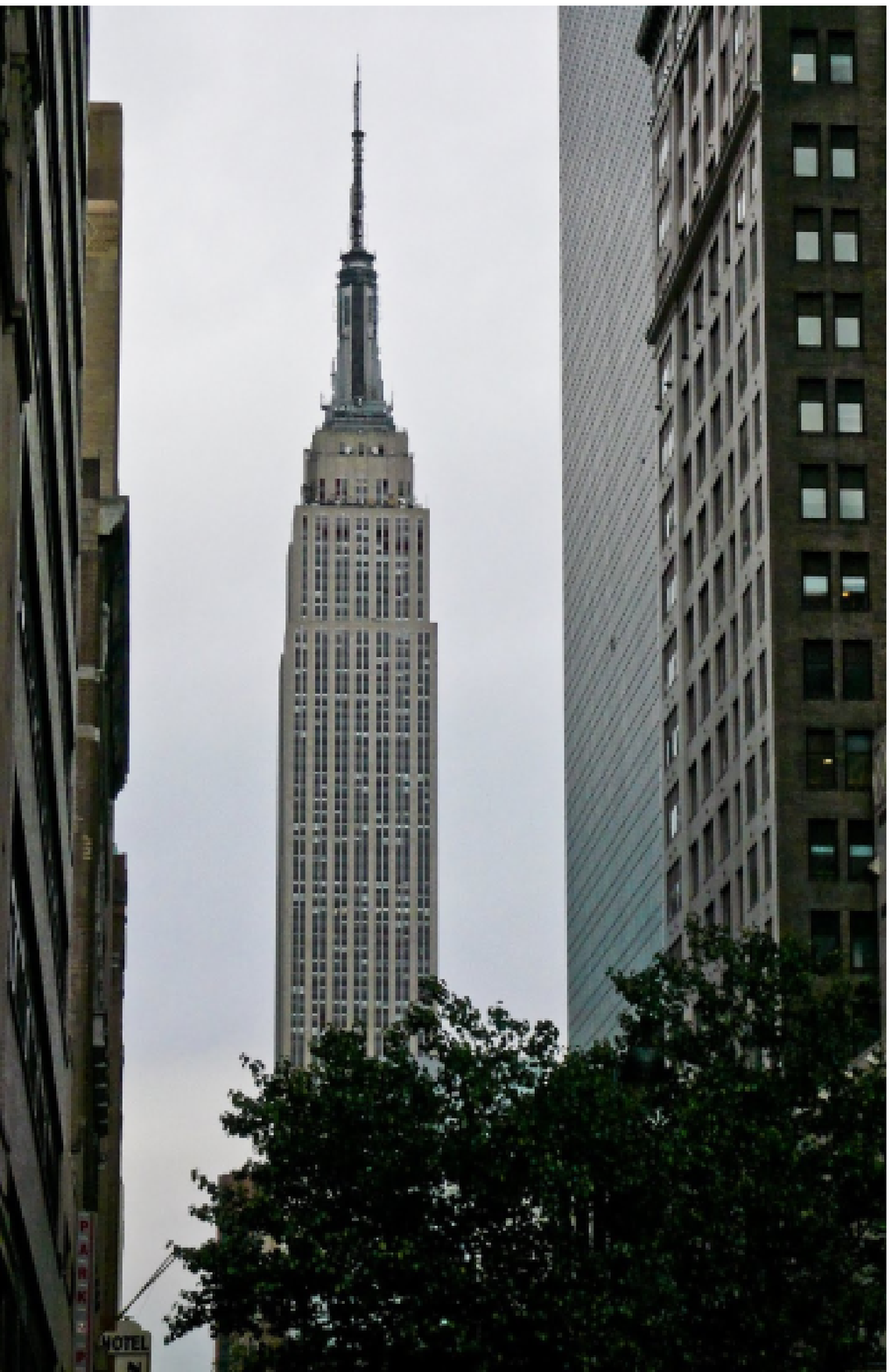} \\
   {\small(a) close = salient} & {\small(b) distant = salient}
   \end{tabular}
\caption{{\bf An ambiguous impact of depth on saliency.} In some cases, the closest object is the salient one (a).
  In other cases, the fact that the object is distant increases its saliency (b).
  }
\label{Fig:depth_ambig}
\end{figure}

We propose a novel Depth-Aware Video Saliency approach that exploits depth information to establish saliency in video sequences (see Figure~\ref{Fig:teaser}).
Integrating depth information as a simple prior into video saliency algorithms (\eg increasing the saliency of close objects) is insufficient due to the ambiguity of depth impact on saliency.
As demonstrated in Figure~\ref{Fig:depth_ambig}, in some cases the closest object attracts the most attention, while in other cases distant objects are the salient ones.

To determine the correct impact of depth on saliency, we train a generative convolutional neural network.
The network predicts a saliency map for a frame, given the map of the previous frame.
This prediction resolves the ambiguity of depth impact by learning its influence on the saliency.

To the best of our knowledge, a comprehensive eye tracking database for video sequences containing depth information is yet to be developed.
To evaluate the performance of our approach we introduce Depth-Aware Video Saliency (DAViS) dataset.
DAViS dataset includes videos that represent the scenarios where depth-aware saliency is beneficial.
The ground-truth was established by recording eye-fixations while viewing the video on regular screens, ignoring the depth information.
Moreover, to establish an objective baseline for the comparison we incorporate depth into the video saliency approach recently proposed by~\cite{rudoy2013Learning}.

Our contribution is threefold.
\begin{itemize}
\vspace{-0.1in}
\item First, we introduce a novel depth-aware video saliency approach and implement it using a generative convolutional neural network.
We show that our approach outperforms state-of-the-art methods for video saliency.
\vspace{-0.1in}
\item  Second, we present a new comprehensive dataset of RGBD videos with eye-fixation ground-truth.
\vspace{-0.1in}
\item Third, we experimentally demonstrate that integrating depth information into saliency estimation framework, which is based on learning, improves its accuracy.
\end{itemize}

The rest of the paper is organized as follows.
Section~\ref{sec:related} reviews the previous work on saliency.
Section~\ref{sec:database} describes our database and the baseline algorithm for depth-aware saliency.
Section~\ref{sec:saliency} introduces our depth-aware video saliency approach.
Section~\ref{sec:results} presents our experimental results.
Section~\ref{sec:conclusion} concludes with a discussion of the work.

\section{Related Work}
\label{sec:related}
Researchers have studied human visual attention for decades.
We refer the readers to~\cite{Borji2012benchmark} for a recent comprehensive comparison on state-of-the-art techniques.
This section discusses two saliency aspects closely related to our research: video saliency and depth-aware saliency.

\vspace{-0.17in} \paragraph{Video Saliency:}
Most existing motion saliency methods are based on image attention models by taking into account simple motion cues.
For instance, Guo \etal~\cite{guo2008spatio} adopt an efficient method based on spectral analysis of the frequencies in the video.
Similarly, Cui \etal~\cite{cui2009temporal} concentrate on motion saliency only by analyzing the Fourier spectrum of the video along X-T and Y-T planes.
Le Meur \etal~\cite{le2007predicting} propose a spatio-temporal computational model, which incorporates several visual features by combining achromatic, chromatic and temporal saliency maps.
Kienzle \etal~\cite{Kienzle2007Spatiotemporal} learn a fixation operator from human eye movements collected under video free-viewing, then learn action classification models.
Mahadevan and Vasconcelos~\cite{Mahadevan2010Spatio} model video patches as dynamic textures, to handle complicated backgrounds and a moving camera.
Seo and Milanfar~\cite{seo2009static} propose using self-resemblance in static and space-time saliency detection.
Rahtu \etal~\cite{Rahtu2010Segmenting} apply a sliding window on video frames comparing the contrast between the feature distribution of consecutive windows.
Kim \etal~\cite{kim2011spatiotemporal} extend the center-surround approach for images to video by adding another dimension.
Mathe \etal~\cite{mathe2012dynamic} present a large scale dataset annotated with human eye movements under various task constraints and show how to train an effective human fixation detector based on this set.
Hou and Zhang~\cite{hou2008dynamic} propose using incremental coding length to measure the rarity of features.
Zhong \etal~\cite{zhong2013video} use optical flows based on the dynamic consistency of motion.
Rudoy \etal~\cite{rudoy2013Learning} narrow their focus to a sparse set of candidate gaze locations and then use learning to predict conditional gaze transitions over time.
Zhou \etal~\cite{Zhou2014TimeMapping} introduce motion saliency method that combines various low-level features with region-based contrast analysis to generate low-frame-rate videos.


\vspace{-0.17in}\paragraph{Depth-Aware Saliency:}
Compared to the number of saliency papers on 2D images and 2D videos, only a small amount of work on 3D content visual attention can be found.
For example,
Jansen \etal~\cite{jansen2009influence} investigate the influence of disparity on viewing behavior in the observation of 2D and 3D still images.
Hakkinen \etal~\cite{hakkinen2010people} examine the difference in the eye movement patterns between viewing of 2D and 3D versions of the same video content.
Liu \etal~\cite{liu2010natural} examine visual features at fixated positions for stereo images with a natural content.
Wang \etal~\cite{wang2012study} examine ``depth-bias'' in the task-free viewing of still stereoscopic synthetic stimuli.
A review of 3D visual attention papers is presented in~\cite{wang2013computational}.

In our research we assume that depth information exists but is not displayed to the viewer.
Thus, we are less concerned about the impact of 3D viewing experience on the human visual perception.
We are interested in exploiting depth for saliency estimation, when the stimuli are two-dimensional.
To the best of our knowledge there is no such previous work for video saliency.

In the domain of still images, integrating depth information into the saliency model was first proposed more than a decade ago by Ouerhani \etal~\cite{ouerhani2000computing}.
They extend the approach of~\cite{itti1998model} and treat depth as just another channel, along with color and other cues.

The recent dramatic improvement of 3D-capable acquisition devices has prompted many researchers to find more effective ways to exploit depth for image saliency calculation.
Ciptadi \etal~\cite{ciptadi2013depth} explicitly construct 3D layout and shape features from the depth measurements.
Niu \etal~\cite{niu2012leveraging} exploit binocular images to estimate a disparity map and only use depth data to identify salient objects.
Lang \etal~\cite{lang2012depth} present a depth prior for saliency learned from human gaze information.
This saliency prior produces a saliency map that is then either directly added or multiplied by the saliency results of other methods.
A novel saliency method, which is based on an anisotropic center-surround difference, is proposed in ~\cite{judepth2014ICIP}.
Desingh \etal~\cite{desingh2013depth} verify that depth really matters on a small dataset and propose to fuse saliency maps, produced by appearance and depth cues independently, through non-linear support vector regression.
Finally, Peng \etal~\cite{peng2014rgbd} propose a saliency model, where  depth and appearance information from multiple layers is taken into account simultaneously, rather than simply fusing depth-induced saliency with color-produced saliency.

\section{Baseline Dataset and Algorithm}
\label{sec:database}
Before presenting our novel depth-aware video saliency approach we discuss a baseline which is required for its evaluation.
Providing a fair performance evaluation of our approach requires the following two components, which are described in the rest of this section:
\begin{enumerate}
\vspace{-0.1in}
    \item dataset of RGBD videos containing ground-truth of human attention
\vspace{-0.1in}
    \item state-of-the-art video saliency estimation algorithm, extended to take into account depth information
\end{enumerate}

\subsection{DAViS Dataset}
An overview of eye-tracking datasets can be found in~\cite{winkler2013overview}.
To evaluate the performance of our approach a comprehensive database containing a ground-truth of human attention on RGBD video sequences is needed.
We are not aware of such a dataset.
Thus, we built a new dataset of RGBD videos and capture human attention when displaying the RGB information only.
We call it the DAViS dataset and we intend to make it publicly available.

\vspace{-0.17in}
\paragraph{Collecting the videos:}
The videos in our dataset should represent the scenarios where depth-aware saliency is beneficial.
Thus, we focus on RGBD videos acquired by built-in phone/tablet/laptop depth/stereo cameras or 3D sensors, such as Kinect or LiDAR.
We consider acquisitions devices that can be either static or installed on moving vehicles or robots.
Thus, we include video sequences of static and dynamic scenes, acquired by static and dynamic sensors, indoors and outdoors.
We cover scenarios such as video conference, surveillance, tracking and obstacle avoidance.

To achieve diversity, we included in the DAViS dataset RGBD videos that were selected from seven publicly available databases~\cite{Karsch2014depth,lai_icra14,Richardt2012RGBZcamera,Silberman12indoor,sturm12iros,xiao2013sun3d,zhou2013dense}.
These datasets were not designed for saliency detection, but rather for other tasks, such as reconstruction, tracking or action recognition.
Thus, they lack the ground-truth of human attention.
We have included only videos where the color and depth frames are fully synchronized.
After ignoring videos with missing regions of the depth map, we included in the DAViS dataset 54 videos with varying durations ranging from 25 to 200 seconds.
The videos were converted to a 30 frame-rate, resulting in approximately 100K frames across all videos.



\vspace{-0.17in}
\paragraph{Building the ground-truth:}
To build a ground-truth for DAViS dataset we conducted a comprehensive user study.
To identify where participants were looking while watching the films, we monitor their eye movements using a \textit{Gazepoint GP3} Eye Tracker.
Video presentation was controlled using the Gazepoint Analysis Standard software.

For the purpose of the study we recruited 91 participants (52 males, 39 females).
Ages ranged from 20 to 67 with the mean age of 26.
All the participants had normal or corrected-to-normal vision and were na\"{\i}ve to the underlying purposes of the experiment.

First, we performed a calibration procedure by asking the participants to look at five red dots appearing on the screen.
Then, we informed the participants that they would watch a series of short videos.
We displayed the videos in random order at a viewing distance varying between 70 and 110 cm.

\begin{figure}[t]
\centering
   \begin{tabular}{cc}
   \includegraphics[width=0.4\linewidth]{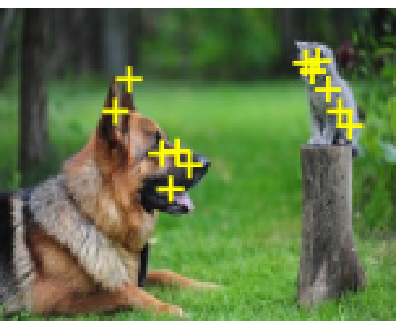} &
   \includegraphics[width=0.4\linewidth]{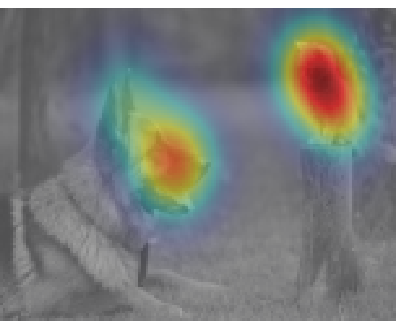} \\
   {\small(a) human fixations} & {\small(b) probability map}
   \end{tabular}
\caption{{\bf Gaze probability map.} Given a sparse ground-truth set of human fixations, marked with yellow '+' per each viewer, we convert it into a dense probability map by convolving with a constant-size Gaussian kernel ($\sigma$ is 5\% of the frame diagonal).}
\label{Fig:db_gaze_preprocess}
\end{figure}

Finally, to get a dense probability map, we convolved the sparse set of fixations from all the participants with a constant-size Gaussian kernel.
Figure~\ref{Fig:db_gaze_preprocess} demonstrates an example of the fixation set and its resulting probability map.

\vspace{-0.17in}
\paragraph{Quality of the ground-truth:}
To assess the quality of the collected ground-truth we quantify the homogeneity of the human fixations.
In other words, we measure how much the fixation map ``explains itself.''
This quality measure also serves as an upper bound for saliency prediction.

To calculate the quality, we randomly divide the set of individual fixation maps, $\textbf{F}$, into two subsets and the probability maps of each subset are compared using $\chi^2$ metric.
We repeat this random process $N$ times and average the results to obtain the homogeneity score for each frame:
\vspace{-0.1in}
\begin{equation}\label{eq:gaze_quality}
    Q =  1 - \frac{1}{N} \sum_{i=1}^{N} \mathlarger {\mathlarger { \chi^2}}  (M(\textbf{F}_i), M(\textbf{F} \setminus \textbf{F}_i)),
\end{equation}
where $\textbf{F}_i \subseteq \textbf{F}$ is a random subset of the fixation set \textbf{F} in the $i-th$ iteration and $M(\textbf{F})$ is the dense probability map of $\textbf{F}$.
The final quality score for each video is calculated by averaging the scores over all the frames.

We compare the quality of our ground-truth to quality of the DIEM (Dynamic Images and Eye Movements) dataset~\cite{mital2011clustering}.
DIEM is a well-known dataset, which has been widely used for evaluation of video saliency techniques.
It includes 84 high-definition videos of varying styles.
The dataset is provided together with gaze tracks of about 50 participants per video.
The video clips included in the DIEM dataset lack any depth information.

\begin{figure}[t]
\centering
   \begin{tabular}{c}
   \includegraphics[width=0.9\linewidth]{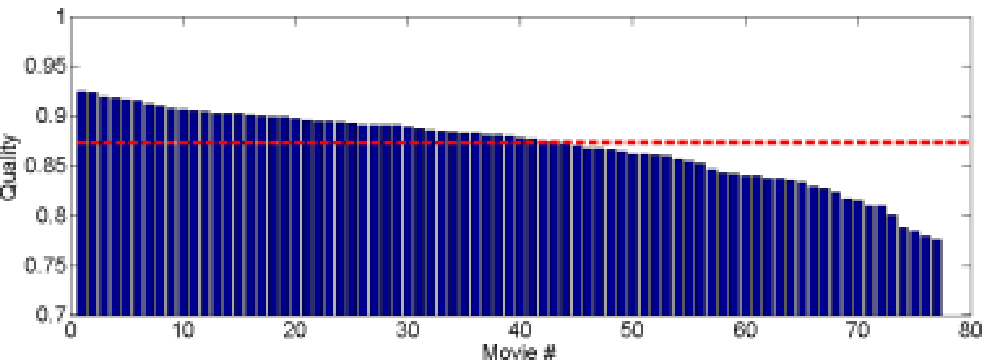} \\
   {\small (a) DIEM dataset~\cite{mital2011clustering} }\\
   \includegraphics[width=0.9\linewidth]{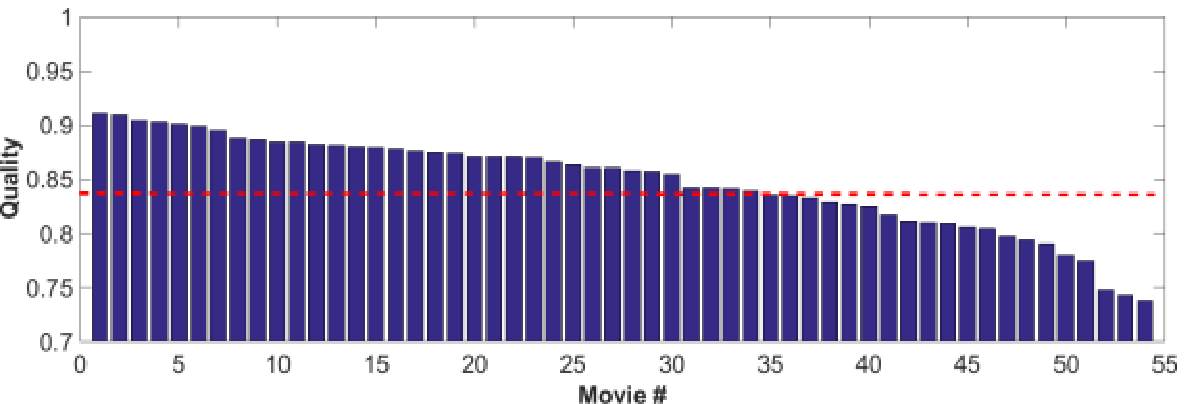} \\
   {\small (b) DAViS dataset}
   \end{tabular}
\caption{{\bf Quality of the gaze ground-truth.} To assess the quality of the collected ground-truth we measure how much the fixation map ``explains itself.''
(Each bar corresponds to one video; the red line indicates the average).
The quality of the fixation maps in our DAViS dataset is comparable to the one of the DIEM dataset~\cite{mital2011clustering}.
}
\label{Fig:db_gaze_quality}
\end{figure}

Figure~\ref{Fig:db_gaze_quality} compares the quality of our gaze tracking ground-truth (DAViS) to the quality of the fixation data in the DIEM dataset.
Perfect correlation, meaning that all the participants followed the exact same focus point on the screen, corresponds to a score of 1.

The DIEM dataset contains movies that have been professionally filmed and usually edited with a goal to attract human attention to specific objects on the screen.
This is especially noticed in commercials and movie trailers.
Thus, we expect high homogeneity of the human fixations.
Indeed, Figure~\ref{Fig:db_gaze_quality}(a) demonstrates the average score of 0.87 varying from 0.78 to 0.93 between the different movies.

DAViS dataset includes mostly unedited clips, filmed either by amateurs or automatically.
As explained above, we have intentionally included such videos since they represent the scenarios in which depth-aware saliency is beneficial.
Thus, our dataset is more ``challenging'' in this regard, and we cannot always expect people to agree on one specific focus of attention.
Still, as shown in Figure~\ref{Fig:db_gaze_quality}(b), the average score of the DAViS dataset is 0.84 varying from 0.74 to 0.91.
These comparable results indicate that most people agree on the same limited number of attention foci, even when the videos were filmed without trying to draw human attention to specific objects.
We also verify visually that the viewers are not looking at a single point most of the time.

We believe that our DAViS dataset represents the wide range of common scenarios where depth-aware saliency is beneficial.
The size of the DAViS dataset (54 videos) was chosen to be similar to the other two most popular datasets for video saliency: DIEM~\cite{mital2011clustering} and CRCNS~\cite{itti2004automatic}, which include 85 and 50 videos, respectively.

\label{sec:baseline_alg}
\subsection{Baseline Depth-Aware Algorithm}
To establish a fair and objective baseline for the comparison we extend the algorithm recently proposed by~\cite{rudoy2013Learning} with depth information in its key stages.
Let us first summarize the original scheme and then explain our extensions.

\vspace{-0.17in} \paragraph{Original scheme:}

\begin{figure}[t]
\centering
    \includegraphics[width=0.81\linewidth]{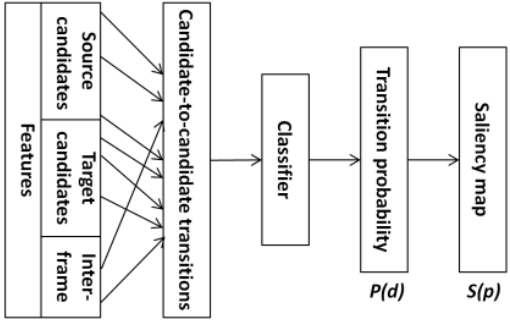}
\caption{{\bf Saliency estimation using explicit transition prediction.}
Initially, the features are calculated on the source candidates (previous frame saliency) and the target ones (detected).
The aggregated features that represent gaze transitions are fed to a trained classifier that outputs a probability of transition to target candidates.
Finally, the probabilities are integrated into a saliency map.}
\label{Fig:alg1}
\end{figure}

As demonstrated in Figure~\ref{Fig:alg1}, first, a sparse set of candidates is generated for each frame.
Then, a classifier that predicts gaze transitions between various candidates of different frames is trained.
The feature space of the classifier accounts for the candidates' properties (\eg saliency magnitude, motion magnitude) and also captures the relation between the candidates (\eg the distance between their locations).
Next, applying the trained classifier, the gaze transition probability from each candidate of a source frame to each candidate of a target frame is calculated.
Finally, a saliency map is generated for each frame based on transition probabilities.

The candidate locations are generated for all video frames based on three cues.
First, Graph-Based Visual Saliency (GBVS)~\cite{harel2007graph} is calculated for each frame.
Second, some high-level cues (\eg human figures and faces) are added to the frame saliency.
Third, to account for motion the optical flow is calculated between the consecutive frames.
Finally, each candidate location is represented by a Gaussian blob, calculated by applying mean-shift clustering and Gaussian fitting on the normalized saliency maps and on the differences in the optical flow magnitude.

After generating a set of candidates in each frame, the gaze transition probability from the candidates of two consecutive frames is calculated.
All the possible pairs of candidates are considered and each pair is associated with a feature vector.
The feature vector consists of (1) the mean saliency of the candidate neighborhood, (2) Difference-of-Gaussians of the optical flow vectors and of their magnitude, (3) discrete candidate labels: face, body and center and (4) geometric features: the distance between the candidates and the distance from the candidate location to the center of the frame.
A classifier is trained on a subset of videos based on the eye-tracking ground-truth.
Finally, the transition probabilities are calculated by applying the classifier on the entire dataset.


\vspace{-0.17in} \paragraph{Depth-aware extension:}
We incorporate depth information in three key stages: static saliency estimation, optical flow calculation and gaze transition modeling.
Our experiments show that all three improvements are vital.

\textbf{First}, depth-aware image saliency is used for generating candidate locations.
We calculate depth-aware saliency based on a multi-stage RGBD model recently proposed in~\cite{peng2014rgbd}.
This technique accounts for both depth and appearance cues derived from low-level feature contrast, mid-level region grouping, and high-level prior enhancement.
\textbf{Second}, depth is used for the optical flow calculation between consecutive frames.
Instead of calculating optical flow on three color channels, we use an additional channel --- the depth.
This way our motion estimation is more accurate than in the previous methods, especially for objects moving on a similarly colored background.
We have considered implementing more sophisticated methods for dense motion estimation using color and depth (\eg~\cite{herbst2013rgb}).
However, the complexity of such techniques is high, making them impractical to apply to videos.
\textbf{Third}, when calculating the feature vectors associated with each candidate pair we exploit depth information, by adding a signed difference in candidates' depths to the set of the geometric features.

All the candidates in the source and destination frames are examined and labeled as positive or negative.
The transitions are positive when they connect between the candidates that are aligned with the human fixations.
Other transitions are marked as negative.

An SVM classifier is trained on the feature vectors and their corresponding labels.
The output of the classifier is the signed distance from the separating hyper-plane.
This distance is proportional to the confidence $C(s,d)$ of transition from the candidate $s$ of the source frame to the candidate $d$ of the destination frame.
The overall probability $P(d)$ of gaze to reach the destination candidate $d$ is calculated by combining all positively classified transitions to candidate $d$.
Thus, ignoring transitions with negative confidence, we calculate $P(d)$ as follows:
\vspace{-0.07in}
\begin{equation}\label{eq:sal_dest}
    P(d) =  \frac{1}{|\mathbf{N}_S|} \sum_{s \in \mathbf{N}_S} S(s) \cdot  \max\big(C(s,d),0\big)  ,
\end{equation}
where $\mathbf{N}_S$ is the set of all the sources and $S(s)$ is the saliency of the source candidate.

Finally, the saliency of pixel $p$ in the destination frame is given by a sum of constant-size Gaussians around each destination candidate $d$, scaled up by the probability $P(d)$:
\vspace{-0.04in}
\begin{equation}\label{eq:sal_final}
    S(p) =  \frac{1}{|\mathbf{N}_D|} \sum_{d \in \mathbf{N}_D} P(d) \cdot \exp\left(-\frac{||p-d||^2}{2\sigma^2}\right),
\end{equation}
where $\mathbf{N}_D$ is the set of all the destination candidates in a given frame and $\sigma$ equals 5\% of the frame diagonal.

\section{Our Approach}
\label{sec:saliency}

\begin{figure*}
\centering
    \includegraphics[width=0.82\linewidth]{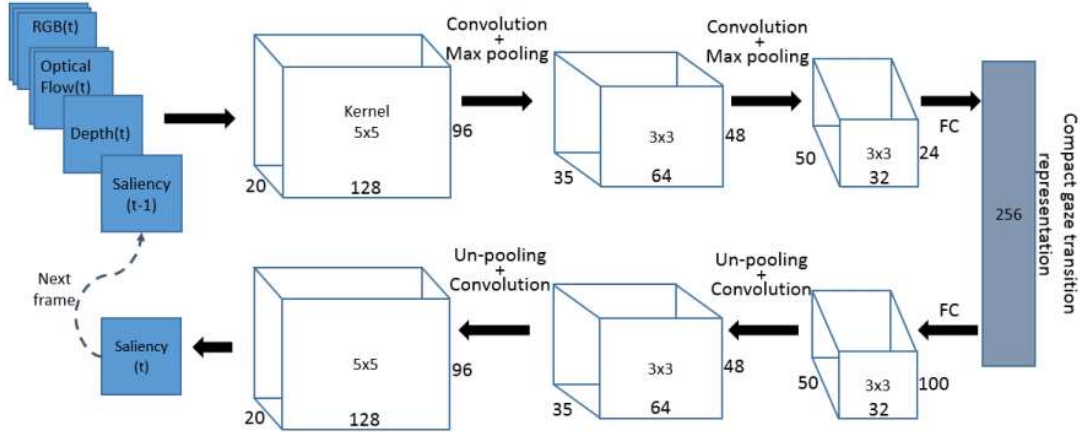}
\caption{{\bf Saliency reconstruction using a generative convolutional neural network.}
The input is the saliency calculated for the previous frame and additional information from the current frame.
Then the data is encoded and only the saliency of the current frame is reconstructed.}
\label{Fig:alg2}
\end{figure*}


This section presents our approach for depth-aware video saliency estimation, which is based on the following three principles.
\textbf{First}, in video the gaze usually slightly varies between frames, and when it does change significantly, it is constrained to a limited number of foci of attention.
\textbf{Second}, people usually follow the action by shifting their gaze to a new interesting location.
Thus, we consider a sparse candidate set of salient locations and use learning to predict transitions between them over time.
\textbf{Third}, in addition to the above two principles, which are common to many previous video saliency approaches, we claim that depth perception has an impact on human attention.
This claim is supported by our experimental results as shown in Section~\ref{sec:results}.
Note that in some cases, the closest object attracts the most attention (Figure~\ref{Fig:depth_ambig}(a)) and in  other cases, a distant object causes humans to concentrate their attention on it (Figure~\ref{Fig:depth_ambig}(b)).
To resolve this ambiguity, we propose incorporating depth into the learning process.

To realize the above three principles, we propose to train a generative convolutional neural network to predict the saliency for each frame.
According to the first and the second principles, the gaze transition between frames is limited to a small number of locations.
Therefore, it is safe to assume that it is feasible to learn a compact representation for gaze transition between frames.
As shown in Figure~\ref{Fig:alg2}, our network's input is the saliency calculated for the previous frame and additional information from the current frame.
Then the data is encoded in a compact way, which represents the gaze transition between frames and only the saliency of the next frame is reconstructed.



Work on generative models typically addresses the problem of unsupervised learning of a compressed, distributed representation (encoding) for a set of data.
Such networks are usually used to generate samples from a hidden representation.
The most known examples are auto-encoders based on Restricted Boltzmann Machines (RBMs)~\cite{hinton2006reducing} and Deep Boltzmann Machines (DBMs)~\cite{salakhutdinov2009deep}.
A basic auto-encoder is an artificial neural network used for learning data coding.
Following the notation from~\cite{masci2011stacked}, first, the input $x$ is mapped to the latent representation $h$ using a function $h = \sigma(W x + b)$.
This compressed representation is then used to reconstruct the input by a reverse mapping of $\hat{x} = \sigma(W'h + b')$.
The weights of $W$ are optimized, minimizing an appropriate cost function over a given training set.
Usually the same weights for encoding the input and the decoding are used, \ie $W'=W^T$.

Conventional auto-encoders are fully connected and consequently ignore the spatial image structure.
This introduces redundancy in the parameters, forcing each feature to be global.
We base our architecture on convolutional auto-encoder structure~\cite{masci2011stacked,baccouche2012spatio}, whose weights are shared among all locations in the input, preserving spatial locality.

In the aforementioned auto-encoders, an input image $x$ is passed through the hidden layers, computing activations at all the layers to obtain the output image $\hat{x}$.
Then, the deviation error from the input $e=x-\hat{x}$ is calculated and back-propagated through the network.

As shown in Figure~\ref{Fig:alg2}, our generative convolutional neural network gets as an input a set of seven images $X=\{x_i\}_1^7$ and reconstructs only one image.
For each frame, the set $X$ consists of the following seven channels: RGB (3 images), optical flow (2 images), depth map and a saliency map $S(t-1)$ calculated for the previous frame.
The output of the network is an estimation of a single saliency map $ \hat{S}(t)$ for the current frame.
Therefore, the deviation error is calculated as $e = S(t) - \hat{S}(t)$, introducing an asymmetry between the input and the output, \ie $W' \neq W^T$.
Then, the error is back-propagated through the network, updating the weights using stochastic gradient descent.
The whole process is recursive, where we start with a saliency map $S(0)$ which consists of a single Gaussian located in the center of the frame.
Then the estimated saliency map $\hat{S}(1)$ is fed as an input $S(1)$ to the network for the next frame.

Finally, following our first principle, we strive to estimate a sparse set of attention foci.
However, the nature of our generative network is to reconstruct relatively smooth output images.
Thus, we add an output post-processing stage to sharpen the peaks of a limited number of attention foci.
This is done by applying mean-shift clustering and Gaussian fitting which results in a mixture of Gaussians.

\vspace{-0.17in} \paragraph{Architecture details:}
We experimented with different network configurations and the best results are achieved by the network shown in Figure~\ref{Fig:alg2}.
First, the input 7-channel image is passed through an encoder.
The encoder consists of three layers of convolutions followed by max-pooling whose sizes are 128x96, 64x48 and 32x24 with kernel sizes of 5x5, 3x3 and 3x3, respectively.
Then the data is encoded in 256 latent variables fully connected to the encoder and the decoder.
The decoder consists of three layers of un-pooling followed by convolution of the same sizes in reverse order.
The un-pooling is performed according to the scheme proposed by~\cite{dosovitskiy2015learning}.

We used stochastic gradient descent with a fixed momentum of 0.9.
For 200 epochs the learning rate was $10^{-4}$ and then for an additional 200 training epochs we divided the rate by two after every 50 epochs.
The network is trained on a subset of two-thirds of the videos and the training error is estimated based on the eye-tracking ground-truth.

Since our saliency learning is recursive, only frames from different videos are used simultaneously, limiting the batch size to the number of videos in the training set.
In other words, the first batch consists of all the first frames, the second batch consists of all the second frames, when the input to the second batch is the saliency maps estimated in the first batch.
For the simplicity of the exposition we used the term ``previous frame''; however in the implementation we use an interval of 10 frames, since it takes up to 10 frames for humans to fixate on a new object.

\section{Results}
\label{sec:results}

This section presents both quantitative and qualitative evaluation of our technique.

\vspace{-0.17in}
\paragraph{Quantitative evaluation:}
For quantitative evaluation we use two common metrics: area-under-curve (AUC) and $\chi^2$ distance between distributions.
AUC is the area under the Receiver Operating Characteristics (ROC) curve~\cite{borji2012exploiting}.
Human fixations are considered as the positive set, while the negative set is formed from randomly sampled points from the image.
The saliency map is then treated as a binary classifier to separate the positive samples from the negative ones.
Thresholding over the saliency map and plotting true positive rate vs. false positive rate results in the ROC curve.

AUC considers the saliency results at the locations of the human fixations.
Thus, it distinguishes purely between a peaky saliency map and a smooth one.
To view the fixations as samples of a distribution, rather than considering each fixation separately, similarly to~\cite{rudoy2013Learning}, we prefer another metric: $\chi^2$ distance between two distributions.
The $\chi^2$ distance prefers a peaky saliency map over a smooth one.

For the $\chi^2$, perfect prediction corresponds to a score of 0.
For AUC, perfect prediction corresponds to a score of 1, while a score of 0.5 indicates the chance level.
Thus, to be consistent, we use $1$ - $\chi^2$ when reporting our results.

To the best of our knowledge, we are the first to propose depth-aware video saliency.
Therefore, to provide a fair evaluation we compare our approach to the extended baseline algorithm (Sec.~\ref{sec:baseline_alg}).
We also compare our approach to video saliency technique~\cite{rudoy2013Learning}, image saliency approach (GBVS)~\cite{harel2007graph}, depth-aware image saliency (RGBD)~\cite{peng2014rgbd} and a Gaussian placed in the center~\cite{judd2010learning}.

\begin{table}
\begin{center}
\begin{tabular}{|l|c|c|}
\hline
Method & $1-\chi^2$  & AUC  \\
\hline\hline
RGBD~\cite{peng2014rgbd} & 0.53 $\pm$ 0.21 & 0.66 $\pm$ 0.19 \\
\hline
GBVS~\cite{harel2007graph} & 0.54 $\pm$ 0.25 & 0.65 $\pm$ 0.21 \\
\hline
Center~\cite{judd2010learning} & 0.56 $\pm$ 0.39 & 0.66 $\pm$ 0.36 \\
\hline
Rudoy\etal~\cite{rudoy2013Learning} & 0.61 $\pm$ 0.26 & 0.68 $\pm$ 0.23 \\
\hline
Extended baseline & 0.64 $\pm$ 0.22 & 0.70 $\pm$ 0.18 \\
\hline
Our approach (w/o depth) & 0.60 $\pm$ 0.23 & 0.68 $\pm$ 0.21 \\
\hline
Our approach (w/ depth) & \textbf{0.70 $\pm$ 0.15} & \textbf{0.75 $\pm$ 0.14} \\
\hline
\hline
Ground-truth & 0.84 $\pm$ 0.05 & 0.88 $\pm$ 0.06 \\
\hline
\end{tabular}
\end{center}
\caption{{\bf Quantitative Evaluation.}
We compare our method to depth-aware image saliency (RGBD)~\cite{peng2014rgbd},
image saliency (GBVS)~\cite{harel2007graph},
a Gaussian placed in the center of the frame (Center)~\cite{judd2010learning},
video saliency (Rudoy \etal)~\cite{rudoy2013Learning} and  the extended baseline algorithm from Section~\ref{sec:baseline_alg}.
The upper bound (Ground-truth) for the saliency prediction is given in Equation~\ref{eq:gaze_quality}.
According to both $\chi^2$ and AUC measures our method is the closest to the ground-truth, outperforming other state-of-the-art methods.
Moreover, it is clear that employing depth in video saliency algorithms which are based on learning improves their accuracy.
}
\label{Tab:results_davis_quant}
\end{table}

Table~\ref{Tab:results_davis_quant} demonstrates a quantitative comparison using two different metrics: $\chi^2$ and AUC.
The ``Ground-truth'' in the bottom row is the upper bound for the saliency prediction, which measures how much the ground-truth fixation map ``explains itself'' (Equation~\ref{eq:gaze_quality}).
We use 38 out of 54 RGBD videos in DAViS dataset for training, while the other 16 videos form the testing set.
To quantify the impact of depth, we also carried out an experiment where we removed the depth information from the input to our approach.

The results of our depth-aware methods are the closest to the ground-truth.
According to both $\chi^2$ and AUC measures, the relative improvement over the state-of-the-art method~\cite{rudoy2013Learning} is approximately 15\% (0.70/0.61).
We also see that employing depth in video saliency algorithms, which are based on learning, improves their accuracy;
both the extended baseline (Sec.~\ref{sec:baseline_alg}) and our approach outperform previous approaches.
Finally, the standard deviation of our approach is lower than in all other methods, making it more reliable than the others.

Note that the trivial approach of a Gaussian, placed in the center of the frame, produces fairly good average results due to two facts.
First, when filming the videos we usually attempt to place the most interesting composition in the center of the frame.
Second, when viewing relatively boring scenes we tend to move the gaze to the center of the frame.
Thus, when comparing this trivial approach to the ground-truth we see relatively a high score in average.
However, the standard deviation of this score is almost twice as high as the standard deviation of other methods, which makes the center-based Gaussian approach highly unreliable.

%

\begin{figure*}[t]
\hspace*{-0.1in}
   \begin{tabular}{cccccc}

%
%
%
%
%
%
%
%

    &{\small Ground-truth}
    &{\small GBVS~\cite{harel2007graph}}
    &{\small Rudoy \etal~\cite{rudoy2013Learning}}
    &{\small Our w/o depth} \\

   \vspace*{-0.17in}
   \multirow{4}{*}{\includegraphics[width=0.31\linewidth]{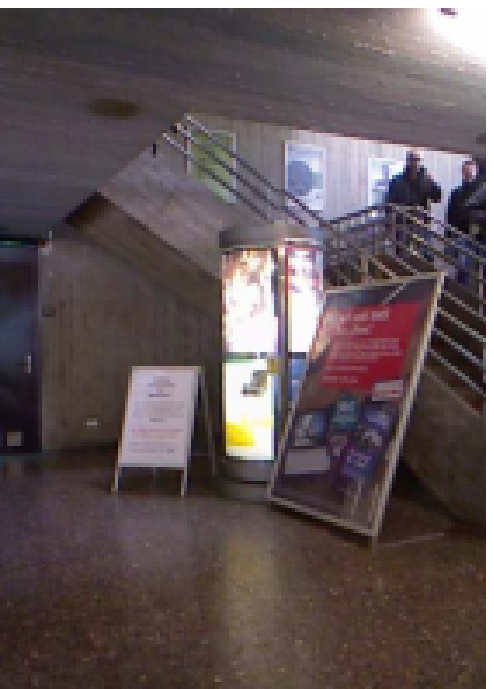} } \\

       &
       \hspace*{-0.1in}
       \includegraphics[width=0.145\linewidth]{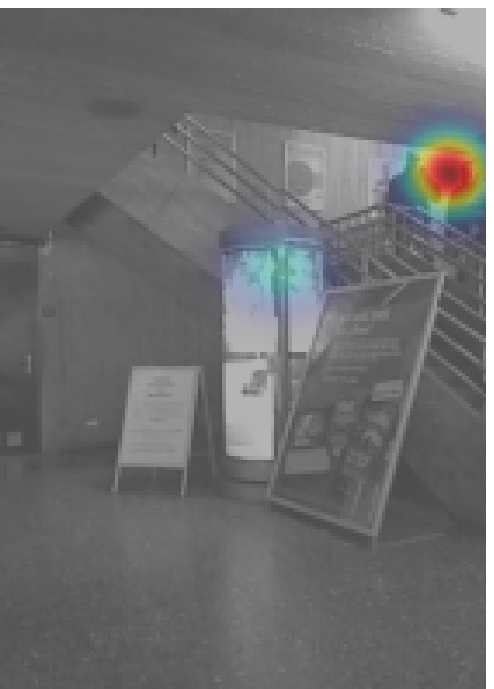} &
       \includegraphics[width=0.145\linewidth]{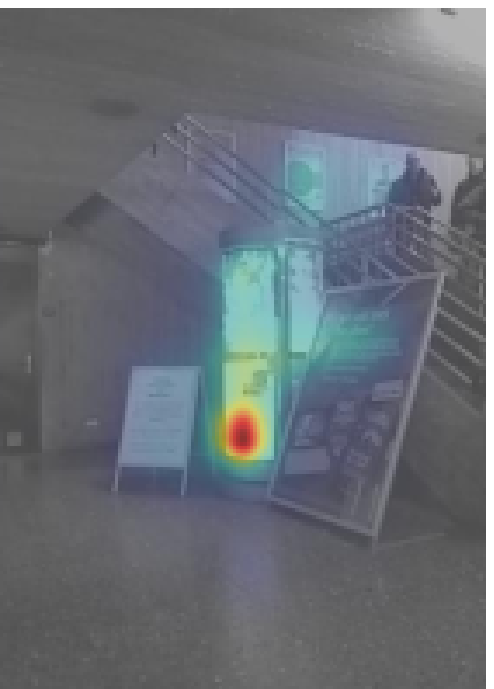} &
       \includegraphics[width=0.145\linewidth]{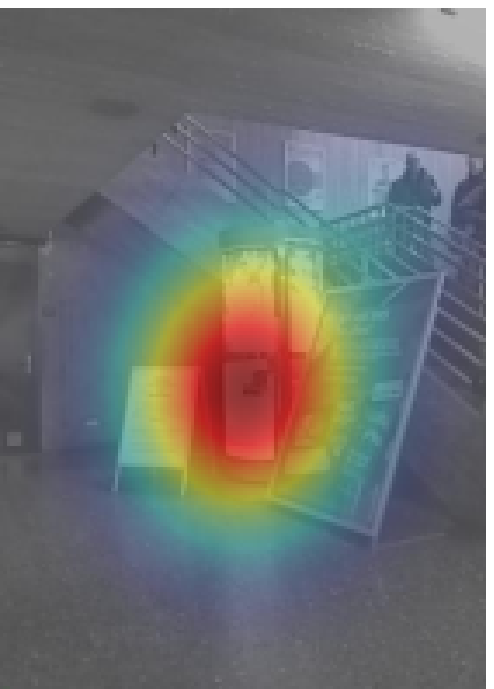} &
       \includegraphics[width=0.145\linewidth]{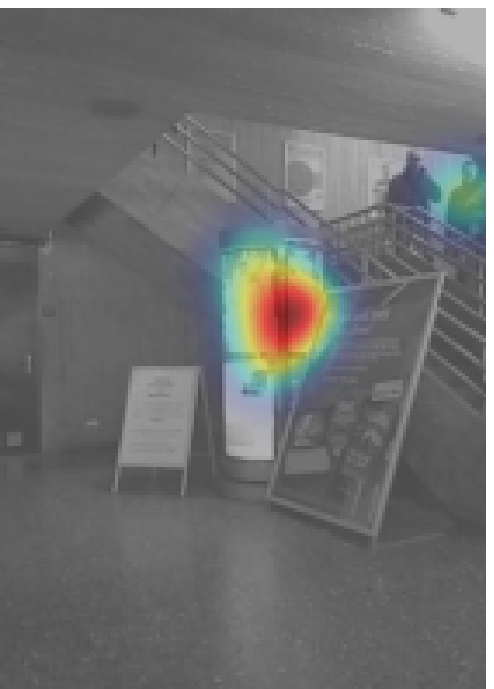} \\
       &{\small Depth}
       &{\small RGBD~\cite{peng2014rgbd}}
       &{\small Extended baseline}
       &{\small Our w/ depth}  \\

       &
       \hspace*{-0.1in}
       \includegraphics[width=0.145\linewidth]{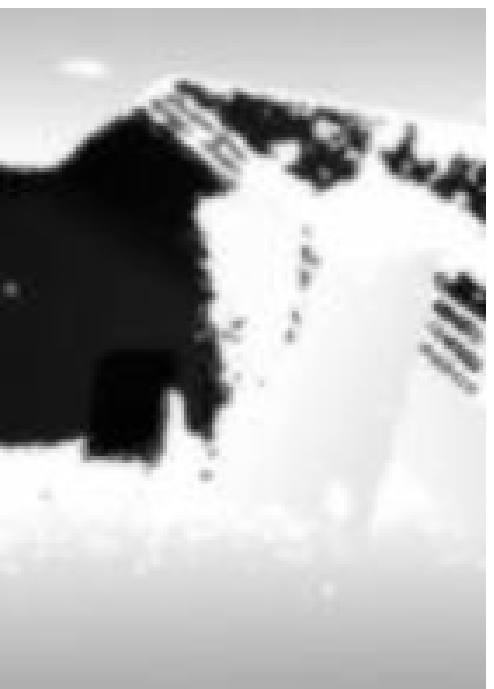} &
       \includegraphics[width=0.145\linewidth]{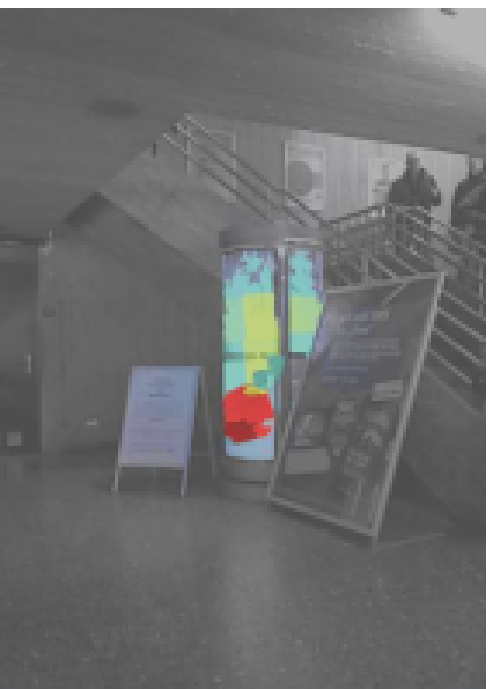} &
       \includegraphics[width=0.145\linewidth]{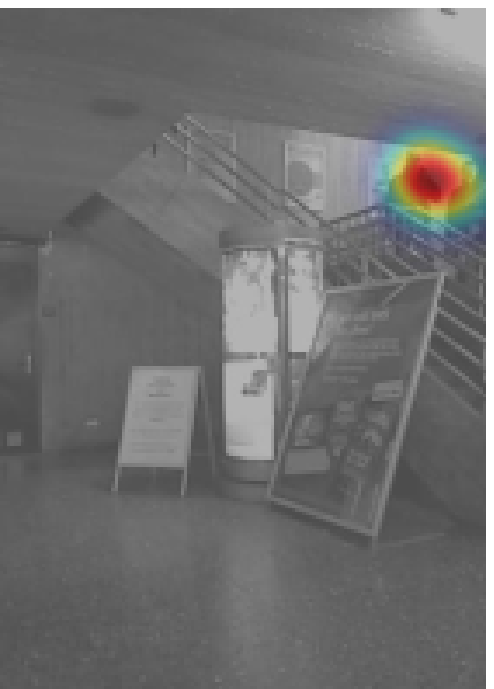} &
       \includegraphics[width=0.145\linewidth]{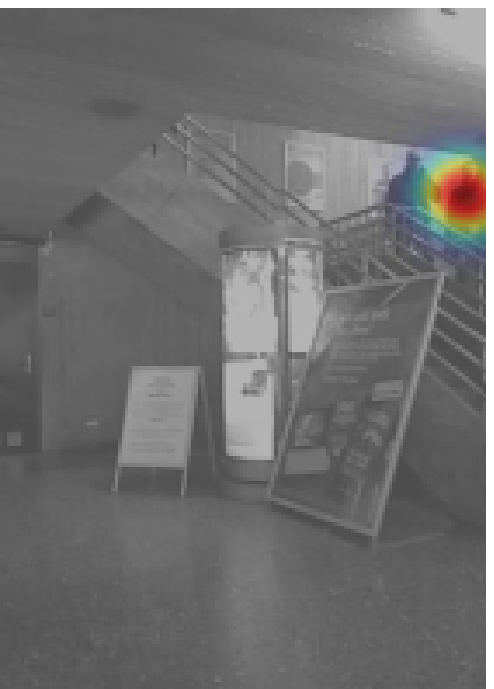} \\

\multicolumn{6}{l}{{\small Depth-aware saliency maps are similar to the ground-truth, detecting interesting motion in the background.}}
\\

    &{\small Ground-truth}
    &{\small GBVS~\cite{harel2007graph}}
    &{\small Rudoy \etal~\cite{rudoy2013Learning}}
    &{\small Our w/o depth} \\

   \vspace*{-0.17in}
   \multirow{4}{*}{\includegraphics[width=0.31\linewidth]{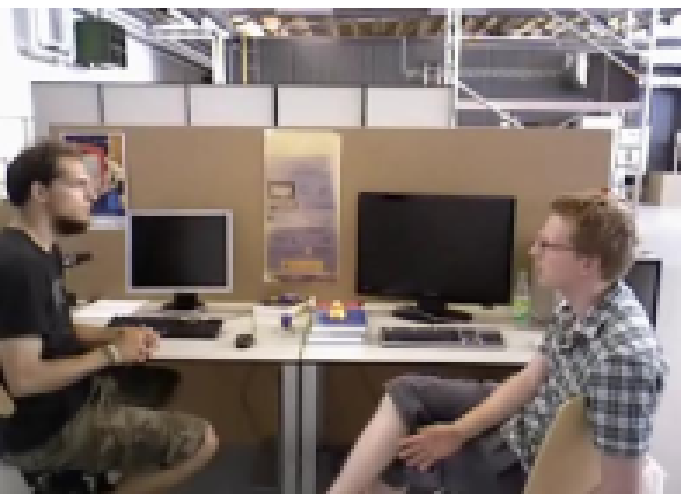} } \\

       &
       \hspace*{-0.14in}
       \includegraphics[width=0.145\linewidth]{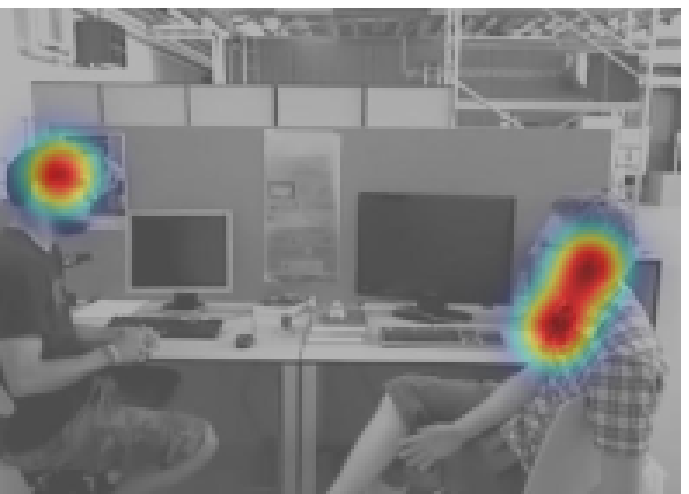} &
       \includegraphics[width=0.145\linewidth]{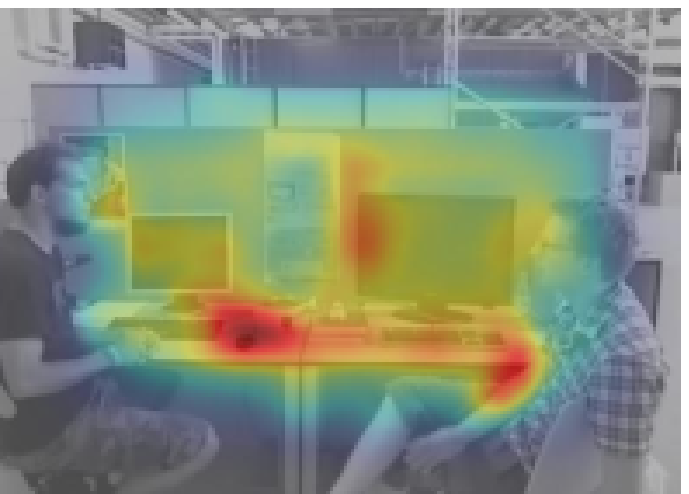} &
       \includegraphics[width=0.145\linewidth]{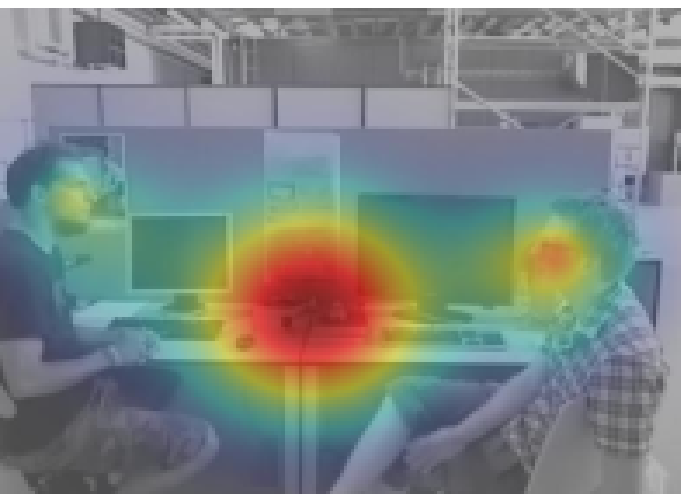} &
       \includegraphics[width=0.145\linewidth]{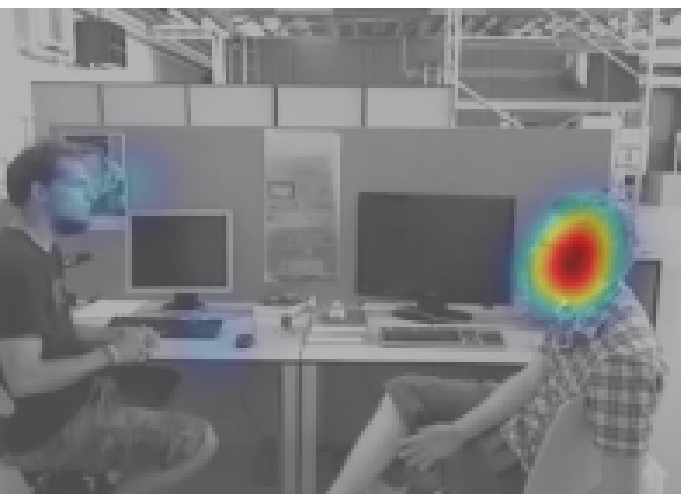} \\
       &{\small Depth}
       &{\small RGBD~\cite{peng2014rgbd}}
       &{\small Extended baseline}
       &{\small Our w/ depth}  \\

       &
       \hspace*{-0.14in}
       \includegraphics[width=0.145\linewidth]{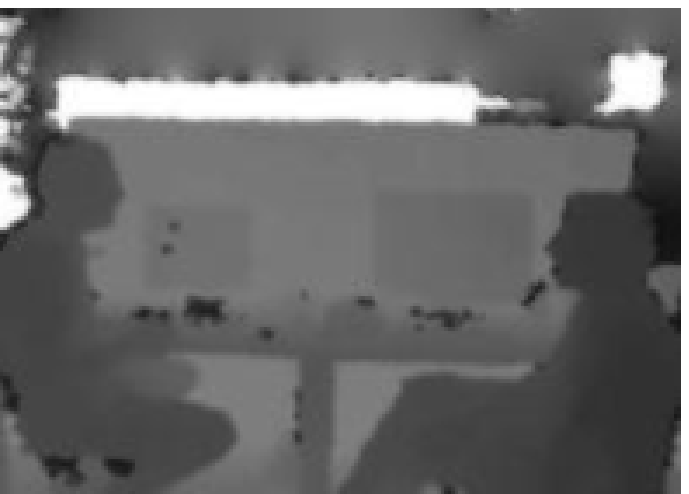} &
       \includegraphics[width=0.145\linewidth]{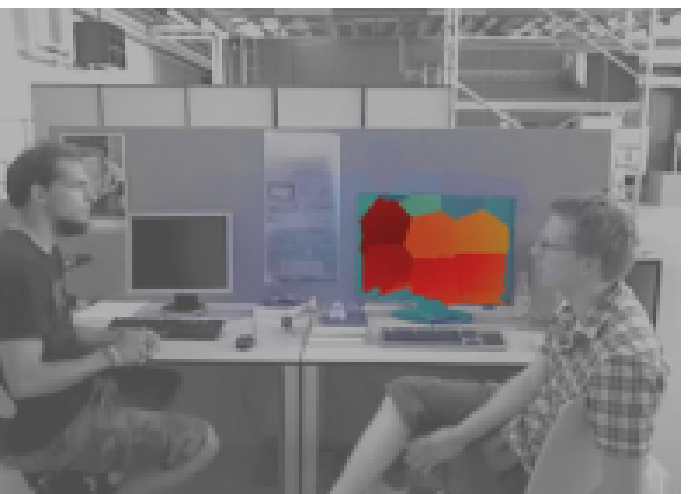} &
       \includegraphics[width=0.145\linewidth]{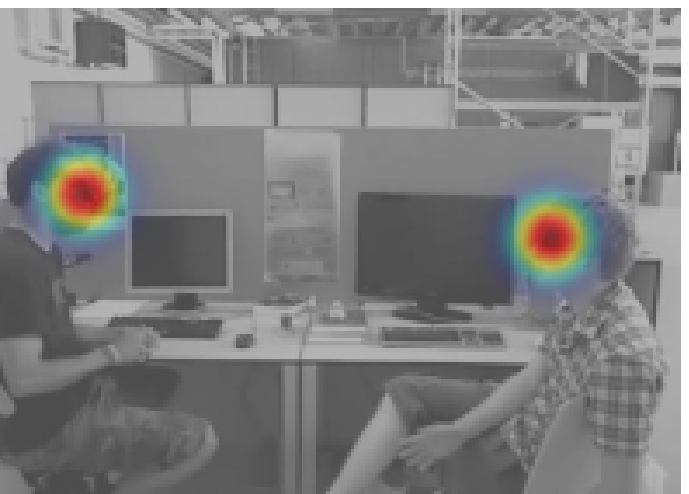} &
       \includegraphics[width=0.145\linewidth]{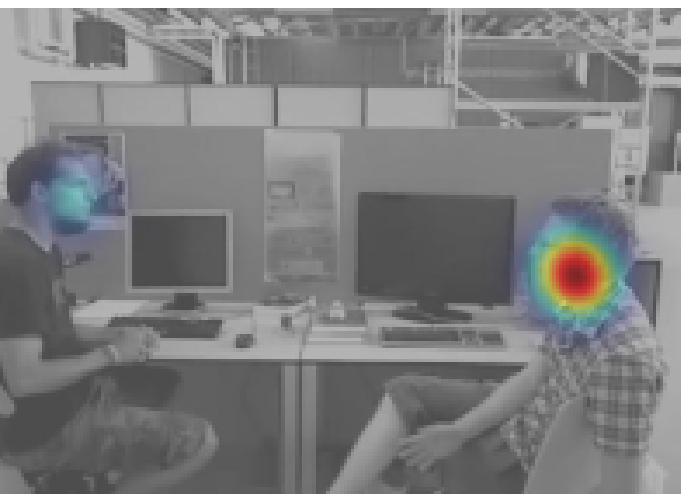} \\

%
%
%
%
%
\multicolumn{6}{l}{{\small While watching a two-person conversation, the gaze shifts from one face the other, which is accurately captured by depth-aware saliency.}}
\\ \\

   \end{tabular}
\caption{{\bf Qualitative evaluation on DAViS dataset.}
We compare our results to the ground-truth and to the additional three saliency methods: video saliency (Rudoy \etal)~\cite{rudoy2013Learning}, image saliency (GBVS)~\cite{harel2007graph} and depth-aware image saliency (RGBD)~\cite{peng2014rgbd}.
The left side of the figure demonstrates the input RGB frame, the depth data and the ground-truth, while the saliency results are shown on the right side.
Both depth-aware methods outperform other state-of-the art methods.
Our novel approach produces more concise results, and this fact is supported by the low standard deviation in Table~\ref{Tab:results_davis_quant}.  }
\label{Fig:results_davis_qual}
\end{figure*}

\vspace{-0.17in}
\paragraph{Qualitative evaluation:}
Figure~\ref{Fig:results_davis_qual} demonstrates a qualitative comparison of our approach to the ground-truth and other saliency techniques.
In both cases the depth-aware saliency map is more visually consistent with the ground-truth than the maps of the other methods.
For example, while watching a conversation between two persons, the gaze shifts from one face to the other, which is accurately captured by depth-aware saliency.The complete video saliency results are given in the supplemental material.

\section{Conclusion}
\label{sec:conclusion}
In this paper, we proposed a novel depth-aware video saliency method, which predicts human foci of attention when viewing 3D video content on 2D screens.
Our method employs a generative convolutional neural network to reconstruct saliency for each frame by implicitly learning the gaze transition from the previous frame.
The network was trained to predict the saliency of the next frame by learning from depth, color, motion and saliency of the current frame.
Experimental results show that exploiting depth is beneficial for video saliency, allowing our method to outperform previously proposed state-of-the-art methods.

Moreover, we presented DAViS dataset, comprehensive dataset of eye-fixation ground-truth for RGBD videos.
DAViS dataset contains videos representing common scenarios where depth-aware saliency is beneficial.
To record eye-fixation ground-truth, we conducted a comprehensive user study, where the RGBD videos were displayed on regular screens ignoring depth information.

We believe that the constructed dataset and our work are helpful to stimulate further research in the area.
In the future we plan to test our methods in various applications, \eg video editing, video compression and video summarization.
We also plan to adapt our technique to stereo videos.




{
\bibliographystyle{ieee}
\small
\bibliography{davis}
}

\end{document}